\documentclass[10pt,twocolumn]{article} 
\usepackage{simpleConference}
\usepackage{times}
\usepackage{graphicx}
\usepackage{amssymb}
\usepackage{url,hyperref}
\usepackage[numbers]{natbib}
\usepackage{amsmath}

\begin{document}

\title{ Locally Adaptive Learning Loss for Semantic Image Segmentation }

\author{Jinjiang Guo$^{1,2}$, Pengyuan Ren$^1$, Aiguo Gu$^1$, Jian Xu$^1$, Weixin Wu$^1$ \\
\\
$^1$Beijing NetPosa Technologies Co., Ltd. Beijing, China \\
$^2$Institut National des Sciences Appliqu\'ees de Lyon, Lyon, France\\
\\
jinjiang.guo@insa-lyon.fr \{renpengyuan, guaiguo, xujian, wuweixin\}@netposa.com \\
}

\maketitle
\thispagestyle{empty}

\begin{abstract}
We propose a novel locally adaptive learning estimator for enhancing the inter- and intra- discriminative capabilities of Deep Neural Networks, which can be used as improved loss layer for semantic image segmentation tasks. Most loss layers  compute pixel-wise cost between feature maps and ground truths, ignoring spatial layouts and interactions between neighboring pixels with same object category, and thus  networks cannot be effectively sensitive to intra-class connections. Stride by stride, our method firstly conducts adaptive pooling filter operating over predicted feature maps, aiming to merge predicted distributions over a small group of neighboring pixels with same category, and then it computes cost between the merged distribution vector and their category label.  Such design can make  groups of neighboring predictions from same category involved into estimations on predicting correctness with respect to their category, and hence train networks to be more sensitive to regional connections between adjacent pixels based on their categories. In the experiments on \emph{Pascal VOC 2012} segmentation datasets, the consistently improved  results show that our proposed approach achieves better segmentation masks against previous counterparts.

\end{abstract}

\section{Introduction}
Convolutional Neural Networks are rapidly driving advances in semantic image segmentation \cite{Everingham2015The,Caesar2016COCO,Cordts2016The,Zhou2017CVPR}, which aims to predict  accurate and effective masks on different classes of targets. To fulfill this challenge, previous study focused on designing or finetuning different network architectures \cite{Jonathan2017Fully,He2017Mask,Girshick2015Fast,Ren2015Faster,Chen2016DeepLab}. To our knowledge,  all these frameworks adapt  the estimator (i.e. loss ) proposed in \cite{Jonathan2017Fully}, which averages pixel-wise cross-entropy over prediction maps and ground truths of input batches. However, this kind of estimator only measures \emph{pixel-wise} distances between predicitons and ground truths, neglecting the interactions between pixels of same category within their neighborhoods. Whereas, such interactions are crucial especially  when the appearances of targets change due to the deformation, illumination variations, occlusion and so forth \cite{Fan2016SANet,Guo2016Subjective}. 

\begin{figure}[t]
  \begin{center}
    \includegraphics[width=3.5in]{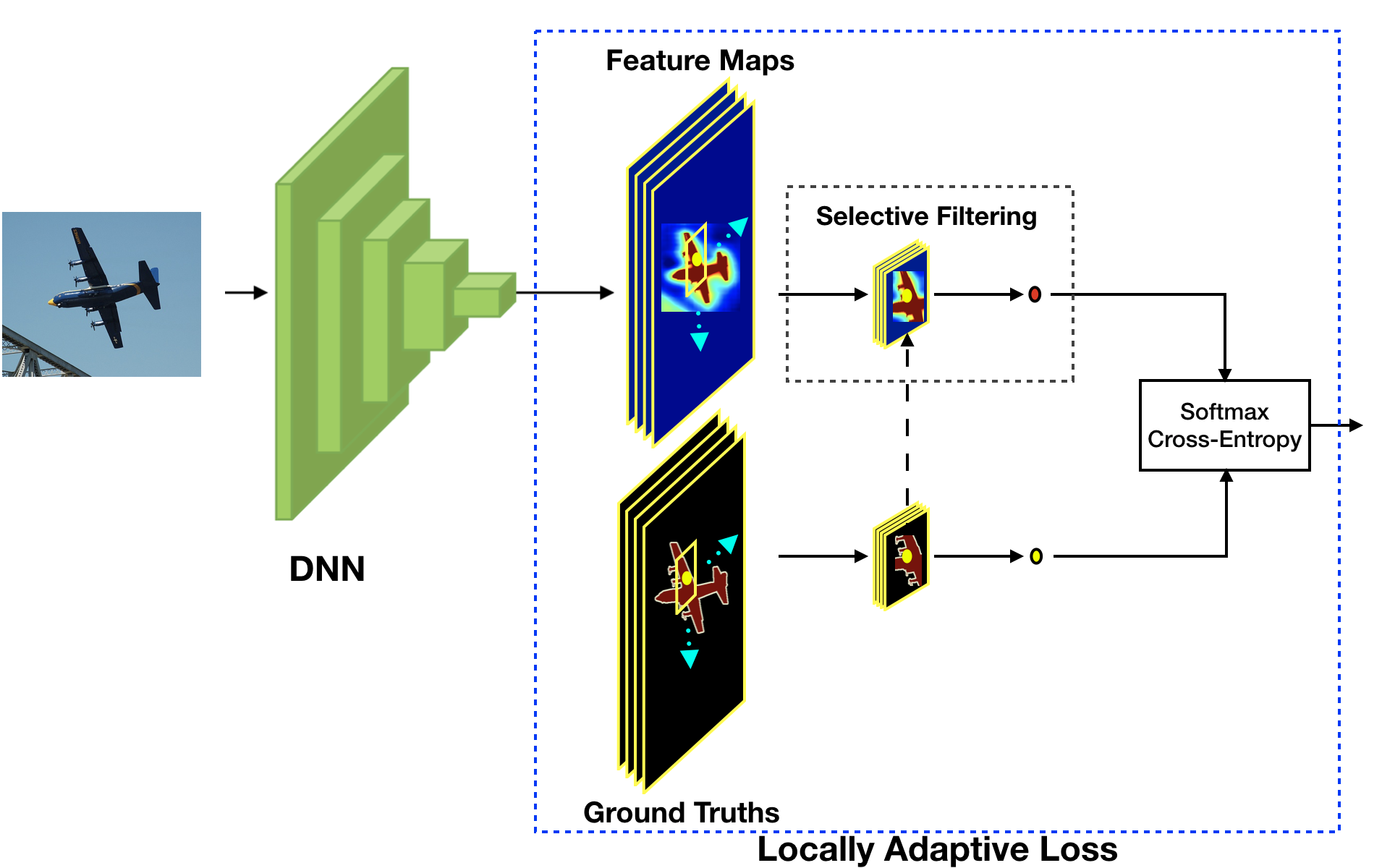}
  \end{center}

  \caption{\small The training framework of our proposed algorithm. The computational filter  slides over both output feature maps and  ground truths simultaneously. At each stride, the predicted vectors of neighboring pixels within filter are selectively pooled into a merged vector based on the label of center pixel. Then, it computes local cost  between the merged vector and center pixel's label. Such operation is conducted on  each valid pixel over input batches, and finally it computes a global loss for each input batch.}
  \label{fig-framework}
\end{figure}

Previous  loss functions for enhancing the intra-class features were designed for image classification  \cite{He2015Deep,Krizhevsky2012ImageNet,Simonyan2014Very,Szegedy2015Going}, which usually measures batch costs  between predicted classes and labels over batchs of images, such like contrastive loss \cite{Hadsell2006Dimensionality,Chen2014Deep}, triplet loss \cite{Schroff2015FaceNet} and center loss \cite{Wen2016A}.   However, as mentioned in \cite{Wen2016A},  the approaches, like contrastive loss and triplet loss, require image pairs or triplets for each training iteration, which result in the dramatic growth of training samples, and thus significantly increase the computational complexity. Center loss overcomes such problem by introducing k-nearest neighbor (k-NN) \cite{Fukunage1975A} algorithms into sfotmax cross-entropy. At each  training iteration, it computes the distances between deep features and every class centers of the features over a mini-batch of images, and updates the centers after each iteration. Center loss can effectively minimize the intra-class variations while keeping the features of different classes separable. Even though, such kind of estimation is still computational expensive, let alone, for image segmentation tasks, each pixel is considered as a training sample. Moreover, most semantic segmentation datasets exhibit long tail distributions with few object categories, which means inter- and intra- classes are imbalanced,  and consequently biasing networks training towards major classes \cite{Bulo2017Loss}.  To address class imbalance problem,  in the realm of object detection, Lin et al. \cite{Lin2017Focal} modified  standard cross entropy loss to down-weight the losses assigned to well-classified examples, and proposed focal loss.

In this paper, we  introduce a novel locally adaptive loss for semantic image segmentation by  estimating selectively filtered predictions based on their categories. Figure \ref{fig-framework} illustrates the training framework of our proposed method at a glance. The selective pooling filter slides over output feature maps and ground truths simultaneously, meanwhile at each striding step, it selectively pools predicted vectors into a merged one, then computes cost between the merged vector and center pixel's category label inside filer. Such operation is conducted on each valid pixel over input batches, and finally it computes a global loss  for each input batch (see Figure \ref{fig-selective}). During training, such loss layer emphasizes on the interactions from same category over neighborhood,  which intuitively indicates that stochastic gradients descent(SGD) solver should optimize  entire predictions on  same category in a scale rather than per pixel.  Such loss  can effectively supervise networks to summarize features of the same category, meanwhile, indirectly enlarge the differences of inter-class features. Thus, the discriminative capabilities of learned models are significantly improved with higher robustness and object sensitivity.  Via this loss, we trained deep neural networks (DNN), and demonstrate that our learned models outperforms against  previous state-of-arts.

In summary, we make the following contributions:
\begin{description}
\item[$\bullet$]  We propose a novel locally adaptive loss layer for  semantic image segmentation. During  learning procedure, it helps networks to improve the capabilities  of discriminating  targets from both inter- and intra- classes.  In our experiments  We also verified  that the learned models trained with our loss outperform against their counterparts.
\item[$\bullet$]  We  explore a simple method for rebalancing losses  from image segmentation datasets, which often exhibit long-tail distribution. Our correction mechanisms can prevent networks from biasing towards majority classes.
\item[$\bullet$]  We  implement  other well-known losses (i.e., center loss and focal loss) for image semantic segmentation tasks as our additional contribution. With these losses, the learned models can also predict decent masks, and thus we use them as our counterparts.
\end{description}

The remainder of this paper has the following structure:  Section \ref{sec2} briefly summarizes related work. Section \ref{sec3} constructs the locally adaptive loss. Section \ref{sec4} illustrates and evaluates our locally adaptive loss via several numerical experiments using different training frameworks. Section \ref{sec5} draws conclusions and proposes direction for future work.

\section{Related Work} \label{sec2}
\subsection{Image Segmentation} \label{sec2sub1}
Semantic image segmentation using  convolutional neural networks or deep neural networks(DNN) has achieved several breakthroughs in recent years \cite{BadrinarayananK15, Chen2016DeepLab, He2017Mask, Jonathan2017Fully,Cordts2016The, DaiLHS16}. Inspired by the work \cite{Jonathan2017Fully}, researchers commonly remove  last fully connected (FC) layers of neural networks,  and then utilize the   in-network upsampled  or deconvolved  predictions of convolutional layers as predicted feature maps. The estimating procedure for training generally  computes pixel-wise losses between the maps and ground truths over each batch, and then pools them into a global value for back propagation (BP).  

The pixel-wise losses in \cite{Jonathan2017Fully,Chen2016DeepLab} are based on softmax  and multinomial cross-entropy between predicted vectors of neurons and labels. However, this computation collapses the spatial dimensions of both predicted maps and labeled images into vectors.
The methods like \cite{Dai2015Instance, Pinheiro2015Learning, Pinheiro2016Learning} resort to FC layers to establish the prediction masks, which requires more complex hyper-parameters. Recently,  He et al. \cite{He2017Mask}  proposed a regional loss computation, using aligned Region of Interest (ROI) \cite{Girshick2015Fast} to maintain each object's spatial layout. On each \emph{aligned ROI}, it conducts a pixel-wise sigmoid and binary loss between predictions and targets labels, eliminating inter-class competition 

Very recently,  Loss Max-Pooling \cite{Bulo2017Loss} was  proposed for handling the imbalanced inter-class datasets of  semantic segmentation. It selectively assigns weights to each pixel  based on their losses, and rebalances datasets by up-weighting  losses contributed by minority classes. 

\subsection{ Weighted Ensemble Entropy Estimator }
Density functions, like cross entropy, are widely used as estimators for training CNN and DNN frameworks (e.g., AlexNet \cite{Krizhevsky2012ImageNet}, the VGG net \cite{Simonyan2014Very},  ResNet \cite{He2015Deep},  DenseNet \cite{Huang2016Densely}, etc.). As discussed in \cite{Sricharan2013Ensemble}, the ensemble of weak estimators can improve the performance of learned models, similar to the methods (e.g., boosting \cite{Schapire1990The}, etc.) proposed in the context of classification. Meanwhile,  a \emph{weighted ensemble entropy estimator} was introduced by optimally combining multiple weak entropy-like estimators(e.g., k-NN entropy functional estimators \cite{Hero2003Asymptotic}, intrinsic dimension estimators \cite{Sricharan2010Empirical}, etc.). The  \emph{weighted ensemble entropy estimator} is defined as:
\begin{equation} \label{eq1}
\hat E_w = \displaystyle\sum_{i=1}^{n} w(i) \hat E_i
\end{equation}

where $\hat E_i$ stands for an individual entropy-like estimator, $n$ means the number of estimators,  and  $w(i)$ is the weight to be optimized, which subject to $\sum_{i\in n}w(i)=1$. It was also verified that such weighted estimator can provide better prediction accuracy and stronger discrimination ability with  higher convergence rate. Note that each weighted weak estimator $\hat E_i$ operates on the same set of input variables $\mathbf X$.  Similarly, center loss \cite{Wen2016A} also estimates on the same set of features. 

In contrast, we explored a new training loss for image segmentation task, which estimates on the merged intra-correlated predictions for neighboring pixels with same category. We demonstrate that our new loss can help  networks to better learn the  interactions of neighbor pixels with same category, and  thus improve the discriminative  abilities on both intra- and inter- classes. Our learned models outperform  their counterparts  trained via plain pixel-wise estimators.

\section{The Locally Adaptive Loss} \label{sec3}
In this section, we elaborate the estimating procedure of our proposed method, and demonstrate that our locally adaptive loss can improve the discriminative power of the learned models, followed by some discussions.

\subsection{Selective Pooling Estimator in Scale} \label{sec3sub1}
In image semantic segmentation, the main task of an effective loss function is to improve the discriminative capability of learned model. However, in contrast to the image detection and classification where each  training batch contains independent samples (e.g., labeled images, bounded objects, etc.), each batch for image segmentation contains all the labeled pixels from different objects, which means several groups of input samples are partially correlated to each other once they belong to same object category. Intuitively, estimating the predictions within a small scale of pixels with same category and minimizing the loss over that scale  give a way.

\begin{figure}[t]
  \begin{center}
    \includegraphics[height=5cm]{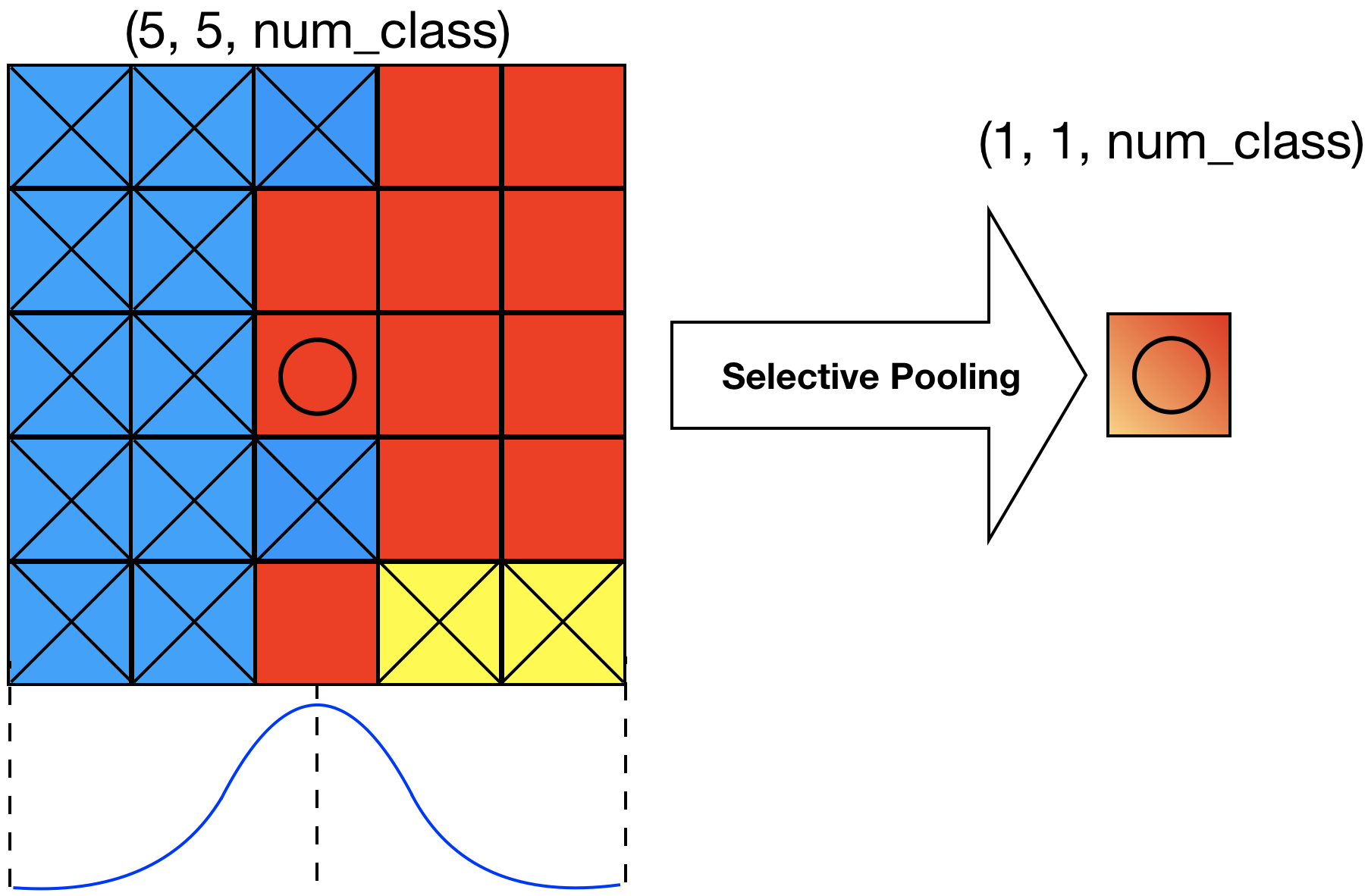}
  \end{center}

  \caption{\small Illustration of  local selective pooling process. The filter (with kernel size $(5, 5, num\_class)$) selects predicted vectors with the same class to center pixel (shown in red color), neglecting predictions from different classes (shown in blue and yellow colors). By applying Gaussian weighting function, it averages all the selected vectors, pooling them into a vector with size of $(1, 1, num\_class)$. Such operation does not increase any predicted vector. }
  \label{fig-selective}
\end{figure}

To this end, at each spatial point (i.e. pixel) $p_{ij}$ with location $(i,j)$ on feature maps, we firstly obtain its \emph{normalized} predicted distribution vector $\mathbf x_{ij} \in \mathbb R^{c \times 1}$. Then  we  conduct our selective pooling  kernel with size $(w, h, c)$, operating on predicted vectors of neighboring points (where $w$, $h$, and $c$ represent: width and height of the kernel window, number of classes respectively). The filtered vector $\mathbf f(i, j) \in \mathbb R^{c \times 1}$ is then computed as follows: 
\begin{equation}\label{equ2}
\mathbf f(i,j) = \frac{1}{\xi} \sum_u \sum_v\mu(i+u, j+v)\omega_d(u,v)\mathbf x_{i+u, j+v}
\end{equation}
\[
    \mu(i+u, j+v)= 
\begin{cases}
    1,& \text{if } \mathbf y_{i+u, j+v} = \mathbf y_{ij}\\
    0,              & \text{otherwise}
\end{cases}
\]
and then our proposed local estimator  is formulated as:

\begin{equation}\label{equ3}
\mathcal L^S(i,j) = \mathcal E(\mathbf f(i,j))
\end{equation}
where $\mathcal E( \cdot )$ represents a local cost function (e.g., softmax cross-entropy, etc.), one-hot label vectors $ \mathbf y_{ij}, \mathbf y_{i+u, j+v}  \in \mathbb R^{c \times 1}$ denotes the $y$th category  of $p_{ij}$ and its neighboring points $p_{i+u, j+v}$. $\mathbf x_{i+u, j+v}  \in \mathbb R^{c \times 1}$  means the normalized  predicted vector at each neighbor point $p_{i+u, j+v}$. $\omega_d(u,v)$ is the Gaussian weighting function based on spatial distance to center $p_{ij}$,  and $\mu(i+u, j+v)$ is a indicator function for eliminating the predictions of neighboring points with different classes from center $p_{ij}$.  $\xi$ is the number of points which have the same labels to $p_{ij}$.  Figure \ref{fig-selective} illustrates the computation details of our selective pooling filter. 

The intuition behind Eqn.\ref{equ3} is that the local estimation computes the entire loss over a group of neighboring points with same category, which indicates the optimization should emphasize on minimizing the overall loss towards a certain category in a scale rather than per point. Note that, for each point, such operation only modifies the distribution of its predicted vector, and does not increase any predicted vector.

For the normalization of  predicted vector $\mathbf x_{ij}$, we use \emph{standard score nomalization}, which is defined as: $ \mathbf x_{ij} =( \hat{\mathbf x_{ij}} -  \mu)/\sigma$, where $\hat{\mathbf x_{ij}}$ means the raw predicted vector, $\mu$ and $\sigma$ stand for mean and standard deviation of  $\hat{\mathbf x_{ij}}$. In practice, we adapt \emph{softmax cross-entropy} as our local cost function  $\mathcal E( \cdot )$. Thus, our local estimator can be rewritten as:

\begin{equation}
\mathcal L^S(i,j) =-\sum_l y_l \log ( \frac{e^{f(i,j)_l}}{\sum_k e^{f(i,j)_k}})
\end{equation}

where $y_l$ represents each element value of center point's label vector $\mathbf y_{ij}$. $f_l$ means each element value of the merged vector $\mathbf f(i,j)$ from the filter.

\subsection{Striding and Batch Pooling Stragety}\label{sec3sub2}
The selective estimator above constructs an ensemble cost based on the category label of center point  inside filter. Then, we propose our Locally ADaptive Loss $\mathcal L^{AD}$ by sliding  such  filter (with striding step $s$) over all the input batches $\mathcal B$ (i.e. images), and averaging all the local estimator values with \emph{Minkowski} pooling. Here we take one image as an input batch, and drop indices $i$ and $j$ for description brevity:\\
\begin{equation} \label{eq3}
\mathcal L^{AD} = (\frac{1}{\mathcal M_p} \displaystyle\sum(\mathcal L^S)^k)^{\frac{1}{k}}
\end{equation}
where $\mathcal M_p $ is the total valid number of input batches. As $k$ increases, more emphasis is allocated to the areas with high loss values. This is intuitively practical  to the image segmentation datasets which often contain imbalanced (or skewed) class distributions (e.g., background and people categories account for the majority of input batches). Consequently, such imbalanced datasets bias networks towards the major classes. Thus, it is more reasonable to increase losses of under-represented categories, which often come from the minority classes. Moreover, such batch pooling strategy could increase the impacts of mispredicted samples of intra-class, which acts as a kind of hard sample mining \cite{Chen2017Beyond,Xiao2017Margin}. However, as mentioned before, our method does not increase any losses number after local cost estimation on each point, therefore the computational time   maintains the same at each iteration.

In practice, we set the striding step $s$ to 1, which means the filter slides  pixel by pixel over input feature maps and ground truths. We use Gaussian-\emph{like} weighting as a neat allocation of $\omega_d$ to each neighbor pixel. Specifically, we allocate down-weights $\omega_d$ to prediction vectors of neighbor pixels, according to their chessboard distances $D8$ \cite{Boroujeni2000Modern} to center pixel (i.e., $\omega_d= 2^1$ to center pixel, $\omega_d=2^0$ to the pixels with $D8=1$, $\omega_d=2^{-1}$ to the pixels with $ D8=2$, $\omega_d=2^{-2}$ to the pixels with $D8=3$, and so forth). 

In order to determine $k$, we tried different $k$ values (see table \ref{tab-kvalue}),  and adapted $k=3$ accordingly.  However, in comparison with the learned model trained by plain softmax cross-entropy (last column), no matter which $k$ values adapted, the models via our locally adaptive loss provide consistently higher Mean IoU values on testing dataset. It means the local selective estimator primarily contributes to the effectiveness of locally adaptive loss, rather than batch pooling strategy.
\begin{table}
\centering
\begin{tabular}{l|c|c|c|c|c}
\hline
\hline 
$k$ value & 1 & 2& 3 & 5 & SoftMax CE\\ \hline
Mean IoU & 70.1 & 70.9 & \textbf{71.2} & 70.6 & 66.8 \\ \hline
\hline
\end{tabular}

\caption{\small Influences of batch pooling strategies with different $k$ values. Mean IoUs (in \%)  on Pascal VOC 2012 segmentation validation data. The training framework is based on ResNet-101\cite{Chen2016DeepLab} and trained after 1,000 iterations using our locally adaptive loss with different $k$ values.}
\label{tab-kvalue}

\end{table}

The derivative of $\mathcal L^{AD}$ w.r.t. the input vector $ \mathbf x_{ij}$, written in an element-wise is as follows ($m$, $n$ stand for  each neighbor point's indices inside filter):
\begin{equation}
\frac{\partial \mathcal L^{AD}}{\partial \mathbf x_{ij}} = \frac{\partial \mathcal L^{AD}}{\partial \mathcal L^S} \cdot \frac{\partial \mathcal L^S}{\partial \mathbf x_{ij}} 
\end{equation}

\begin{equation}
\frac{\partial L^{AD}}{\partial \mathcal L^S}  = (\frac{1}{\mathcal M_p})^\frac{1}{k} ( \displaystyle\sum (\mathcal L^S)^k)^{\frac{1}{k}-1}  \displaystyle\sum (\mathcal L^S)^{k-1}
\end{equation}

\begin{equation}
\begin{split}
& \frac{\partial \mathcal L^S}{\partial \mathbf x_{ij}}= \frac{\partial \mathcal L^S}{\partial \mathbf f(i,j)} \cdot \frac{\partial \mathbf f(i,j)}{\partial \mathbf x_{ij}} \\
&  =   \sum_m \sum_n ({\scriptstyle \frac{e^{f(i+m, j+n)_l}}{\sum_k e^{f(i+m, j+n)_k}} -y_{i+m, j+n} )  \mu(i,j) \omega_d(-m, -n) }
\end{split}
\end{equation}

\begin{figure*}[t]
\centering
  \includegraphics[width=0.9\textwidth]{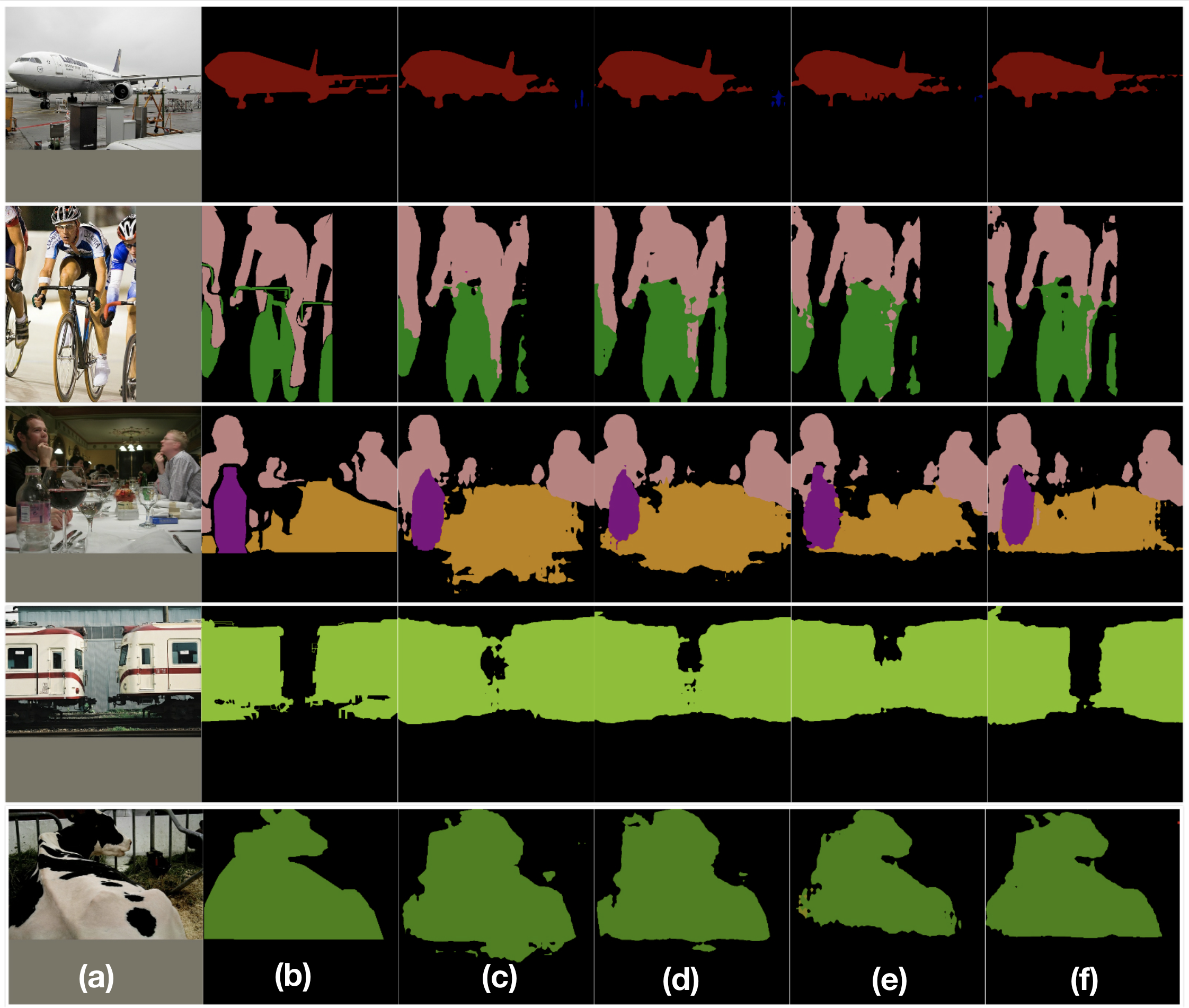}
  \caption{\small Examples of segmented images predicted by learned models with different training losses. Figures \emph{from left to right} represent: (a) original images, (b) ground truths, (c) plain softmax cross-entropy, (d) center loss, (e) focal loss and (f) our locally adaptive loss. With our training loss, the learned model outperforms by predicting more accurate and effective masks. }
    \label{fig-vis}
\end{figure*}

\subsection{Relationship of Locally Adaptive Loss with counterparts}
Both locally adaptive loss and loss max-pooling methods are designed for image semantic segmentation. Locally adaptive loss directly focuses on connections of adjacent pixels with same category, while loss max-pooling aims to rebalance the datasets between majority  and minority classes.

Mostly, locally adaptive loss is a metric approach in the feature space (i.e., activations of last upsampled DNN layer), using the selective pooling filter to increase network attentions on ensemble predicting correctness of neighboring pixels. It applies the filtering operations before local cost computations with ground truths. For loss max-pooling, it only re-weights losses after local cost computations, aiming to increase the contributions of under-represented object classes.

Loss max-pooling is in some way similar to our  batch cost pooling strategy, as we use simple \emph{Minkowski} pooling for handling the imbalanced class datasets. Loss max-pooling can be also embedded into our loss as a replacement for batch pooling strategy.

Loss max-pooling is essentially  a weighting and sampling method on outputs of cost function, which can be considered as hard sample mining \cite{Chen2017Beyond,Xiao2017Margin}, whereas locally adaptive loss operates on both inputs and outputs.

\section{Experiments} \label{sec4}
We have evaluated our novel locally adaptive training loss ($\mathcal L^{AD}$) on the extended  \emph{Pascal VOC} \cite{Everingham2010The} semantic image segmentation datasets. We adapt Intersection-over-Union (IoU) as the evaluating metric on over all classes of datasets. 

\subsection{Network Arichitecture}
For all experiments, we applied DNN network DeepLabV2 proposed in \cite{Chen2016DeepLab}, and implemented with TensorFlow \cite{Abadi2016TensorFlow}, using cuDNN for improving performance. The GPUs used for our experiments are GeForce TX 1070 and Titan Xp. Specifically, we adapted a fully-convolutional ResNet-101 \cite{He2015Deep}  with atrous extensions \cite{Holschneider1989A,Yu2015Multi} for  base layers before adding atrous spatial pyramid pooling (ASPP) \cite{Chen2016DeepLab}. For our baseline method, we applied upscaling (i.e., deconvolution layers with learned weights) before  the softmax cross-entropy (SoftMax CE). In our experiments, the baseline method gives similar results in \cite{Chen2016DeepLab}. Besides baseline method, we also applied center loss \cite{Wen2016A} and focal loss \cite{Lin2017Focal} as replacements of softmax cross-entropy,  to study the effectiveness of losses from different research areas. We also disabled both multi-scale input to networks and post-operations with conditional random fields (CRF), so that we can precisely conclude on our proposed method without complementary. However, all these  complementary methods, including loss max-pooling, can be integrated into our method in case of   demanding better overall performance, when given adequate  devices and training time. For most, our primary purpose is to demonstrate the effectiveness of locally adaptive loss, competing with other baselines  under comparable parameters. We only report results obtained from a single DNN trained using the stochastic gradient descent (SGD) solver, where  we set the initial learning rate to $2.5 \times 10^{-4}$, both decay rate and momentum to 0.9. For data augmentation, we adapted random scale perturbations in the range of $ [0.5,   1.5]$, and horizontal flipping of images.

\begin{figure}
\centering
  \includegraphics[width=0.5\textwidth]{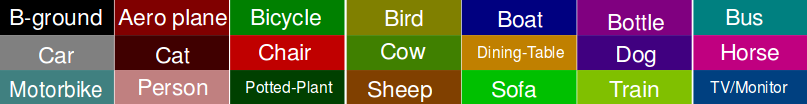}
  \caption{\small Color scheme of   \emph{Pascal VOC} \cite{Everingham2010The}  segmentation datasets.}
    \label{fig-color}
\end{figure}

\subsection{Experiment Results}
We evaluated the performance of our loss $\mathcal L^{AD}$ on the extended  \emph{Pascal VOC} \cite{Everingham2010The}  segmentation benchmark datasets, which consist of 20 object categories and a background category (see Figure \ref{fig-color}).  We set batch size to 2 with crop size of $(321,321)$, using training set with 10,582 images and testing set with 1,449 images. For hyper parameters of our loss, we used kernel sizes of $(3, 3, 21)$, $(5, 5, 21)$ and $(7, 7, 21)$ for our local selective filters, and  ran a total of 20,000 training iterations respectively. For center loss and focal loss, since they were originally designed for object detection and classification, we manually adjusted their hyper parameters to be fitted for image segmentation.  Thus, we set  $\alpha = 0.5$, $\lambda = 3 \times 10^{-4}$ for center loss, and $\alpha = 0.25$, $\gamma = 2$ for focal loss.  We report the mean IoU values in table \ref{results} after 20,000 iterations.  As shown in table  \ref{results}, the learned models trained via locally adaptive loss predict consistently improved results, in particular,  using $\mathcal L^{AD}$ with kernel size: $(5, 5, 21)$ alone can give $1.5\%$ more in predicting accuracy, compared with plain softmax cross-entropy. In contrast,  applying center loss leads to mean IoU values similar to plain softmax cross-entropy, while we obtain $4.2\%$ lower values by using focal loss.

Additionally, in Figure \ref{fig-vis} we exhibit  several segmented examples on testing set to visually demonstrate  improvements  via our training framework against others. The first two columns show the original images (randomly cropped) and ground truths. From the 3rd to 6th columns, we can observe the masks predicted by learned models trained via plain somftmax cross entropy, center loss, focal loss and our locally adaptive loss (with kernel size: $(5, 5, 21)$) respectively. And we can see  that the models via our training framework predict  more accurate and effective masks with higher robustness and object sensitivity, compared to its counterparts.

\begin{table}[h]
\centering
\begin{tabular}{l|c|c}
\hline
\hline
{Training Framework} & {Hyper Para.} & {M IoU} \\ \hline
\small{RN-101 + SoftMax CE}          & -   & 74.6     \\
\small{RN-101 + Center Loss }        & \small{$\alpha = 0.5$, $\lambda = 3 \times 10^{-4}$ } & 74.5     \\
\small{RN-101 + Focal Loss }        &\small{ $\alpha = 0.25$, $\gamma =2$  } & 70.4     \\
\small{RN-101 + $\mathcal L^{AD}$ (ours)}   &\small{ \emph{kernel size:} (3, 3, 21) }& 75.7  \\
\small{RN-101 + $\mathcal L^{AD}$ (ours)  }  &\small{\emph{kernel size:} (5, 5, 21)}   & \textbf{76.1}  \\   
\small{RN-101 + $\mathcal L^{AD}$ (ours)  } &\small{\emph{kernel size:} (7, 7, 21)}   & 75.1  \\      
\hline
\hline     
\end{tabular}
\caption{\small Experimental  results (in \%)  on Pascal VOC 2012 segmentation validation data. The training framework is based on ResNet-101\cite{Chen2016DeepLab}.}
\label{results}
\end{table}

\section{Conclusions and Future Work} \label{sec5}
In this work, we introduced a novel approach to increase networks discriminative capabilities of inter- and intra- class for semantic image segmentations. At each pixel's position our method firstly conducts adaptive pooling filter operating over predicted feature maps, aiming to merge predicted distributions over a small group of neighboring pixels with same category, and then computes cost between the merged distribution vector and their category label. Our locally adaptive loss does not increase any loss numbers, thus the time complexity maintains the same  at each iteration. In the experiments on \emph{Pascal VOC 2012} segmentation datasets, the consistently improved  results show that our proposed approach achieves more accurate and effective segmentation masks against its counterparts. More extensive experiments will be launched on Cityscapes dataset \cite{Cordts2016The} and COCO dataset \cite{Lin2014Microsoft} to further verify our training framework.

\bibliographystyle{plainnat}
\bibliography{refs}

\begin{thebibliography}{40}
\providecommand{\natexlab}[1]{#1}
\providecommand{\url}[1]{\texttt{#1}}
\expandafter\ifx\csname urlstyle\endcsname\relax
  \providecommand{\doi}[1]{doi: #1}\else
  \providecommand{\doi}{doi: \begingroup \urlstyle{rm}\Url}\fi

\bibitem[Abadi et~al.(2016)Abadi, Agarwal, Barham, Brevdo, Chen, Citro,
  Corrado, Davis, Dean, and Devin]{Abadi2016TensorFlow}
Martin Abadi, Ashish Agarwal, Paul Barham, Eugene Brevdo, Zhifeng Chen, Craig
  Citro, Greg~S Corrado, Andy Davis, Jeffrey Dean, and Matthieu Devin.
\newblock Tensorflow: Large-scale machine learning on heterogeneous distributed
  systems.
\newblock 2016.

\bibitem[Badrinarayanan et~al.(2015)Badrinarayanan, Kendall, and
  Cipolla]{BadrinarayananK15}
Vijay Badrinarayanan, Alex Kendall, and Roberto Cipolla.
\newblock Segnet: {A} deep convolutional encoder-decoder architecture for image
  segmentation.
\newblock \emph{CoRR}, abs/1511.00561, 2015.

\bibitem[Boroujeni and Shahabadi(2000)]{Boroujeni2000Modern}
Mehdi~Ahmadi Boroujeni and Mahmoud Shahabadi.
\newblock Modern mathematical methods for physicists and engineers.
\newblock \emph{Measurement Science \& Technology}, 12\penalty0 (12):\penalty0
  2211, 2000.

\bibitem[Bulo et~al.(2017)Bulo, Neuhold, and Kontschieder]{Bulo2017Loss}
Samuel~Rota Bulo, Gerhard Neuhold, and Peter Kontschieder.
\newblock Loss max-pooling for semantic image segmentation.
\newblock 2017.

\bibitem[Caesar et~al.(2016)Caesar, Uijlings, and Ferrari]{Caesar2016COCO}
Holger Caesar, Jasper Uijlings, and Vittorio Ferrari.
\newblock Coco-stuff: Thing and stuff classes in context.
\newblock 2016.

\bibitem[Chen et~al.(2016)Chen, Papandreou, Kokkinos, Murphy, and
  Yuille]{Chen2016DeepLab}
Liang~Chieh Chen, George Papandreou, Iasonas Kokkinos, Kevin Murphy, and
  Alan~L. Yuille.
\newblock Deeplab: Semantic image segmentation with deep convolutional nets,
  atrous convolution, and fully connected crfs.
\newblock \emph{IEEE Transactions on Pattern Analysis \& Machine Intelligence},
  PP\penalty0 (99):\penalty0 1--1, 2016.

\bibitem[Chen et~al.(2017)Chen, Chen, Zhang, and Huang]{Chen2017Beyond}
Weihua Chen, Xiaotang Chen, Jianguo Zhang, and Kaiqi Huang.
\newblock Beyond triplet loss: A deep quadruplet network for person
  re-identification.
\newblock pages 1320--1329, 2017.

\bibitem[Chen et~al.(2014)Chen, Chen, Wang, and Tang]{Chen2014Deep}
Yuheng Chen, Yuheng Chen, Xiaogang Wang, and Xiaoou Tang.
\newblock Deep learning face representation by joint
  identification-verification.
\newblock In \emph{International Conference on Neural Information Processing
  Systems}, pages 1988--1996, 2014.

\bibitem[Cordts et~al.(2016)Cordts, Omran, Ramos, Rehfeld, Enzweiler, Benenson,
  Franke, Roth, and Schiele]{Cordts2016The}
Marius Cordts, Mohamed Omran, Sebastian Ramos, Timo Rehfeld, Markus Enzweiler,
  Rodrigo Benenson, Uwe Franke, Stefan Roth, and Bernt Schiele.
\newblock The cityscapes dataset for semantic urban scene understanding.
\newblock In \emph{Computer Vision and Pattern Recognition}, pages 3213--3223,
  2016.

\bibitem[Dai et~al.(2015)Dai, He, and Sun]{Dai2015Instance}
Jifeng Dai, Kaiming He, and Jian Sun.
\newblock Instance-aware semantic segmentation via multi-task network cascades.
\newblock pages 3150--3158, 2015.

\bibitem[Dai et~al.(2016)Dai, Li, He, and Sun]{DaiLHS16}
Jifeng Dai, Yi~Li, Kaiming He, and Jian Sun.
\newblock {R-FCN:} object detection via region-based fully convolutional
  networks.
\newblock \emph{CoRR}, abs/1605.06409, 2016.

\bibitem[Everingham et~al.(2010)Everingham, Gool, Williams, Winn, and
  Zisserman]{Everingham2010The}
Mark Everingham, Luc~Van Gool, Christopher K.~I. Williams, John Winn, and
  Andrew Zisserman.
\newblock The pascal visual object classes (voc) challenge.
\newblock \emph{International Journal of Computer Vision}, 88\penalty0
  (2):\penalty0 303--338, 2010.

\bibitem[Everingham et~al.(2015)Everingham, Eslami, Gool, Williams, Winn, and
  Zisserman]{Everingham2015The}
Mark Everingham, S.~M.~Ali Eslami, Luc~Van Gool, Christopher K.~I. Williams,
  John Winn, and Andrew Zisserman.
\newblock The pascal visual object classes challenge: A retrospective.
\newblock \emph{International Journal of Computer Vision}, 111\penalty0
  (1):\penalty0 98--136, 2015.

\bibitem[Fan and Ling(2016)]{Fan2016SANet}
Heng Fan and Haibin Ling.
\newblock Sanet: Structure-aware network for visual tracking.
\newblock 2016.

\bibitem[Fukunage and Narendra(1975)]{Fukunage1975A}
K~Fukunage and P.~M Narendra.
\newblock A branch and bound algorithm for computing k-nearest neighbors.
\newblock \emph{IEEE Transactions on Computers}, C-24\penalty0 (7):\penalty0
  750--753, 1975.

\bibitem[Girshick(2015)]{Girshick2015Fast}
Ross Girshick.
\newblock Fast r-cnn.
\newblock \emph{Computer Science}, 2015.

\bibitem[Guo et~al.(2016)Guo, Vidal, Cheng, Basu, Baskurt, and
  Lavoue]{Guo2016Subjective}
Jinjiang Guo, Vincent Vidal, Irene Cheng, Anup Basu, Atilla Baskurt, and
  Guillaume Lavoue.
\newblock Subjective and objective visual quality assessment of textured 3d
  meshes.
\newblock \emph{ACM Trans. Appl. Percept.}, 14\penalty0 (2), October 2016.

\bibitem[Hadsell et~al.(2006)Hadsell, Chopra, and
  Lecun]{Hadsell2006Dimensionality}
Raia Hadsell, Sumit Chopra, and Yann Lecun.
\newblock Dimensionality reduction by learning an invariant mapping.
\newblock In \emph{IEEE Computer Society Conference on Computer Vision and
  Pattern Recognition}, pages 1735--1742, 2006.

\bibitem[He et~al.(2015)He, Zhang, Ren, and Sun]{He2015Deep}
Kaiming He, Xiangyu Zhang, Shaoqing Ren, and Jian Sun.
\newblock Deep residual learning for image recognition.
\newblock pages 770--778, 2015.

\bibitem[He et~al.(2017)He, Gkioxari, Dollár, and Girshick]{He2017Mask}
Kaiming He, Georgia Gkioxari, Piotr Dollár, and Ross Girshick.
\newblock Mask r-cnn.
\newblock 2017.

\bibitem[Hero et~al.(2003)Hero, Costa, and Ma]{Hero2003Asymptotic}
Alfred~O. Hero, Jose~A. Costa, and Bing Ma.
\newblock Asymptotic relations between minimal graphs and alpha-entropy.
\newblock 2003.

\bibitem[Holschneider et~al.(1989)Holschneider, Kronland-Martinet, Morlet, and
  Tchamitchian]{Holschneider1989A}
M.~Holschneider, R.~Kronland-Martinet, J.~Morlet, and Ph. Tchamitchian.
\newblock A real-time algorithm for signal analysis with the help of the
  wavelet transform.
\newblock pages 286--297, 1989.

\bibitem[Huang et~al.(2016)Huang, Liu, van~der Maaten, and
  Weinberger]{Huang2016Densely}
Gao Huang, Zhuang Liu, Laurens van~der Maaten, and Kilian~Q. Weinberger.
\newblock Densely connected convolutional networks.
\newblock 2016.

\bibitem[Krizhevsky et~al.(2012)Krizhevsky, Sutskever, and
  Hinton]{Krizhevsky2012ImageNet}
Alex Krizhevsky, Ilya Sutskever, and Geoffrey~E. Hinton.
\newblock Imagenet classification with deep convolutional neural networks.
\newblock In \emph{International Conference on Neural Information Processing
  Systems}, pages 1097--1105, 2012.

\bibitem[Lin et~al.(2014)Lin, Maire, Belongie, Hays, Perona, Ramanan, Dollár,
  and Zitnick]{Lin2014Microsoft}
Tsung~Yi Lin, Michael Maire, Serge Belongie, James Hays, Pietro Perona, Deva
  Ramanan, Piotr Dollár, and C.~Lawrence Zitnick.
\newblock Microsoft coco: Common objects in context.
\newblock 8693:\penalty0 740--755, 2014.

\bibitem[Lin et~al.(2017)Lin, Goyal, Girshick, He, and Dollar]{Lin2017Focal}
Tsung~Yi Lin, Priya Goyal, Ross Girshick, Kaiming He, and Piotr Dollar.
\newblock Focal loss for dense object detection.
\newblock pages 2999--3007, 2017.

\bibitem[Long et~al.(2015)Long, Shelhamer, and Darrell]{Jonathan2017Fully}
Jonathan Long, Evan Shelhamer, and Trevor Darrell.
\newblock Fully convolutional networks for semantic segmentation.
\newblock \emph{IEEE Transactions on Pattern Analysis and Machine
  Intelligence}, 39\penalty0 (4):\penalty0 640--651, 2015.

\bibitem[Pinheiro et~al.(2015)Pinheiro, Collobert, Doll, and
  Piotr]{Pinheiro2015Learning}
Pedro~O Pinheiro, Ronan Collobert, Doll, and R~Piotr.
\newblock Learning to segment object candidates.
\newblock pages 1990--1998, 2015.

\bibitem[Pinheiro et~al.(2016)Pinheiro, Lin, Collobert, and
  Dollár]{Pinheiro2016Learning}
Pedro~O. Pinheiro, Tsung~Yi Lin, Ronan Collobert, and Piotr Dollár.
\newblock Learning to refine object segments.
\newblock pages 75--91, 2016.

\bibitem[Ren et~al.(2015)Ren, He, Girshick, and Sun]{Ren2015Faster}
Shaoqing Ren, Kaiming He, Ross Girshick, and Jian Sun.
\newblock Faster r-cnn: towards real-time object detection with region proposal
  networks.
\newblock In \emph{International Conference on Neural Information Processing
  Systems}, pages 91--99, 2015.

\bibitem[Schapire(1990)]{Schapire1990The}
Robert~E Schapire.
\newblock The strength of weak learnability.
\newblock \emph{Machine Learning}, 5\penalty0 (2):\penalty0 197--227, 1990.

\bibitem[Schroff et~al.(2015)Schroff, Kalenichenko, and
  Philbin]{Schroff2015FaceNet}
Florian Schroff, Dmitry Kalenichenko, and James Philbin.
\newblock Facenet: A unified embedding for face recognition and clustering.
\newblock In \emph{IEEE Conference on Computer Vision and Pattern Recognition},
  pages 815--823, 2015.

\bibitem[Simonyan and Zisserman(2014)]{Simonyan2014Very}
Karen Simonyan and Andrew Zisserman.
\newblock Very deep convolutional networks for large-scale image recognition.
\newblock \emph{Computer Science}, 2014.

\bibitem[Sricharan et~al.(2013)Sricharan, Wei, and Rd]{Sricharan2013Ensemble}
K~Sricharan, D.~Wei, and Hero~Ao Rd.
\newblock Ensemble estimators for multivariate entropy estimation.
\newblock \emph{IEEE Transactions on Information Theory}, 59\penalty0
  (7):\penalty0 4374--4388, 2013.

\bibitem[Sricharan et~al.(2010)Sricharan, Raich, and
  Iii]{Sricharan2010Empirical}
Kumar Sricharan, Raviv Raich, and Alfred O.~Hero Iii.
\newblock Empirical estimation of entropy functionals with confidence.
\newblock \emph{Statistics}, 2010.

\bibitem[Szegedy et~al.(2015)Szegedy, Liu, Jia, Sermanet, Reed, Anguelov,
  Erhan, Vanhoucke, and Rabinovich]{Szegedy2015Going}
Christian Szegedy, Wei Liu, Yangqing Jia, Pierre Sermanet, Scott Reed, Dragomir
  Anguelov, Dumitru Erhan, Vincent Vanhoucke, and Andrew Rabinovich.
\newblock Going deeper with convolutions.
\newblock In \emph{Computer Vision and Pattern Recognition}, pages 1--9, 2015.

\bibitem[Wen et~al.(2016)Wen, Zhang, Li, and Qiao]{Wen2016A}
Yandong Wen, Kaipeng Zhang, Zhifeng Li, and Yu~Qiao.
\newblock \emph{A Discriminative Feature Learning Approach for Deep Face
  Recognition}.
\newblock Springer International Publishing, 2016.

\bibitem[Xiao et~al.(2017)Xiao, Luo, and Zhang]{Xiao2017Margin}
Qiqi Xiao, Hao Luo, and Chi Zhang.
\newblock Margin sample mining loss: A deep learning based method for person
  re-identification.
\newblock 2017.

\bibitem[Yu and Koltun(2015)]{Yu2015Multi}
Fisher Yu and Vladlen Koltun.
\newblock Multi-scale context aggregation by dilated convolutions.
\newblock 2015.

\bibitem[Zhou et~al.(2017)Zhou, Zhao, Puig, Fidler, Barriuso, and
  Torralba]{Zhou2017CVPR}
Bolei Zhou, Hang Zhao, Xavier Puig, Sanja Fidler, Adela Barriuso, and Antonio
  Torralba.
\newblock Scene parsing through ade20k dataset.
\newblock In \emph{The IEEE Conference on Computer Vision and Pattern
  Recognition (CVPR)}, July 2017.

\end{thebibliography}

\end{document}